\pdfoutput=1

\documentclass[11pt]{article}

\usepackage{acl}

\usepackage{times}
\usepackage{latexsym}

\usepackage[T1]{fontenc}

\usepackage[utf8]{inputenc}
\DeclareUnicodeCharacter{FF09}{)}
\DeclareUnicodeCharacter{03A3}{$\Sigma$}
\usepackage{microtype}

\usepackage{inconsolata}
\usepackage{booktabs}
\usepackage{tabularx}
\usepackage{listings}

\usepackage{amssymb}
\usepackage{tikz}
\usetikzlibrary{calc, backgrounds, positioning}
\usetikzlibrary{decorations.pathreplacing}
\usepackage{amsmath}
\usepackage{stmaryrd}
\usepackage{url}
\usepackage{multirow}
\usepackage{paralist}
\usepackage{color}
\usepackage{caption}
\usepackage{subcaption}
\usepackage{hyphenat}
\usepackage{fvextra}
\lstdefinestyle{promptbox}{
  basicstyle=\ttfamily\scriptsize,
  breaklines=true,
  columns=fullflexible,
  showstringspaces=false,
  tabsize=2,
  numbers=none
}
\usepackage{verbatim}
\usepackage{array}
\title{Teaching LLM to be Persuasive: Reward-Enhanced Policy Optimization for Alignment from Heterogeneous Rewards}

\author{Xia Zeng\thanks{Corresponding author: \texttt{x.zeng@qmul.ac.uk}}, 
        Yihan Chen, Luhui Liu, Chao Luo, Ye Chen, \and Zhuoran Zhuang \\
        Fliggy Alibaba \\
        \texttt{x.zeng@qmul.ac.uk, cyh\_dlut@163.com, luhuiliuse@gmail.com,} \\
        \texttt{guotai.lc@alibaba-inc.com, chen.yechen@alibaba-inc.com,} \\
        \texttt{nantian@alibaba-inc.com}}

\begin{document}

\maketitle

\begin{abstract}
We deploy large language models (LLMs) as business development (BD) agents for persuasive price negotiation in online travel agencies (OTAs). The agent must follow a multi-stage Standard Operating Procedure (SOP) and strict guardrails (no over-promising and no hallucinations), while remaining human-like and effective over long, multi-turn dialogues.
We propose Reward-Enhanced Policy Optimization (REPO), a reinforcement learning post-training method that combines heterogeneous rewards: a preference-trained reward model (RM), an LLM-as-a-judge (RJ) for nuanced behaviors (e.g., emotional value and SOP compliance), and rule-based reward functions (RF) (mainly regex-based) for deterministic checks on numerics, formatting, and guardrails.

In expert consensus evaluation (three human experts; 30 online conversations and 45 curated bad cases), REPO improves average dialogue rating to 4.63 (+0.33 over GRPO) and raises the share of conversations with at least one excellent response to 66.67\% (+23.34 pp over GRPO), while achieving a 93.33\% bad-case fix rate with 75.56\% clean fixes.
In a production A/B test on 9{,}653 real customer conversations (vs.\ an intent-driven dialogue system), REPO improves response rate by +12.14 pp and task success rate by +5.94 pp ($p<0.001$).
\end{abstract}

\begin{figure*}[t]
    \centering
    \includegraphics[width=\textwidth]{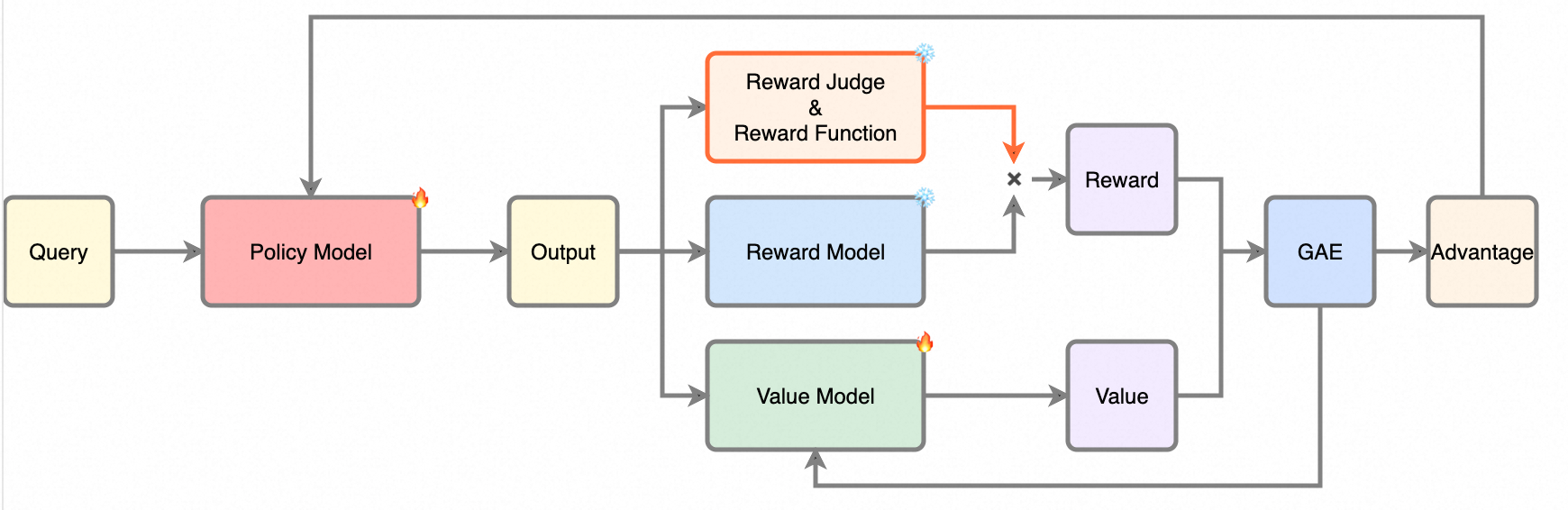}
\caption{Framework of REPO for Alignment from Heterogeneous Reward Signals. Given a query (\(q\)), the Policy Model generates an output (\(o\)), which is evaluated through three reward components: the Reward Model (RM), providing a dense, human preference-aligned signal; the Reward Judge (RJ), an LLM-based evaluator scoring nuanced, high-level behaviors; and rule-based Reward Functions (RF) (mainly regex-based), validating task-specific requirements with deterministic checks. These signals are combined to compute the total reward (\(r\)), used to guide learning via Generalized Advantage Estimation (GAE) to calculate the Advantage (\(A\)). The Value Model predicts the state value (\(v\)) to further refine the training process.}
    \label{fig:REPO}
\end{figure*}

\begin{figure}[htb]
    \centering
    \includegraphics[scale=0.14]{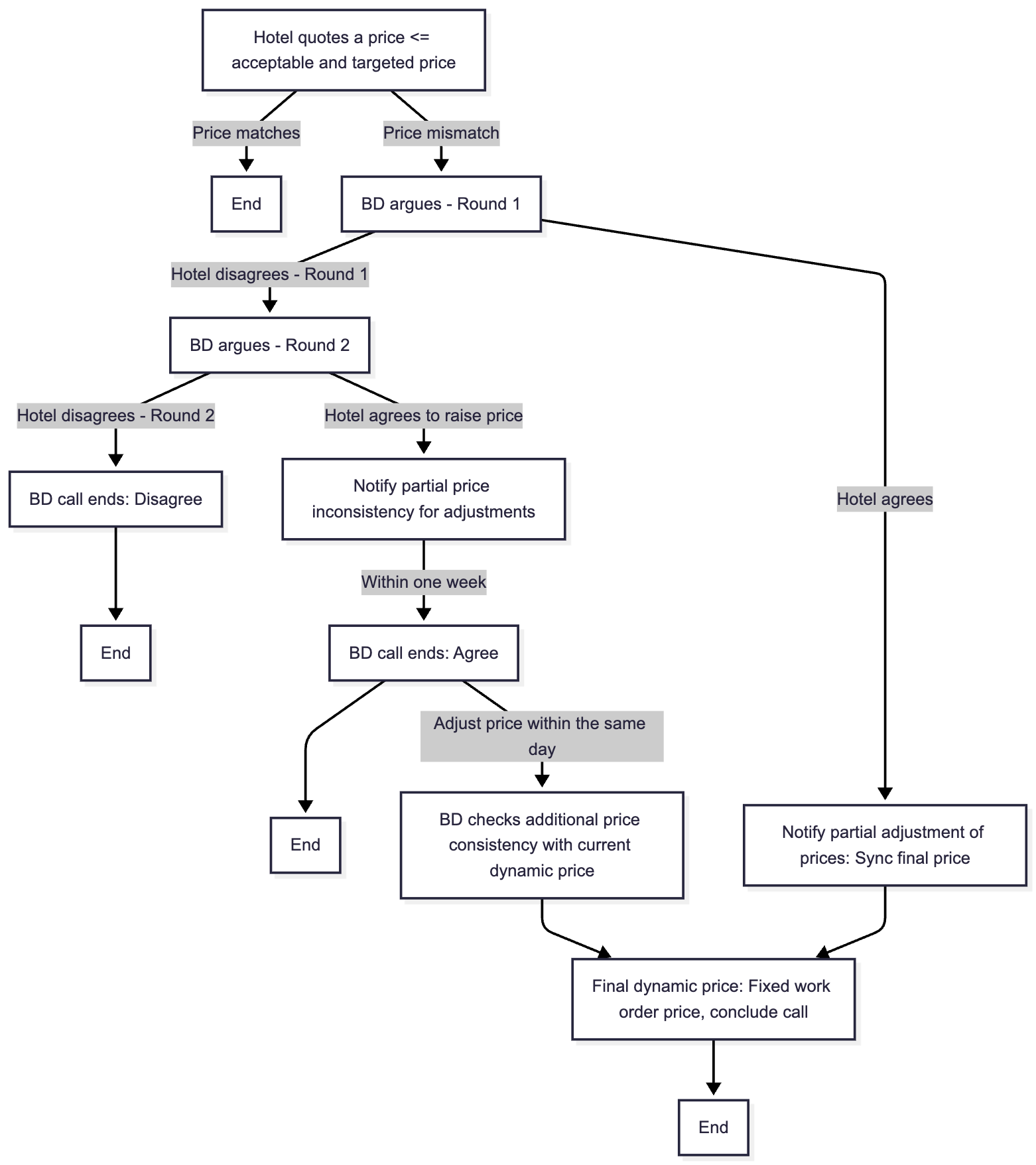}
    \caption{Simplified SOP for the Price Negotiation Task.}
    \label{fig:sop}
\end{figure}

\section{Introduction}
Proactive price negotiation is a core operational workflow in online travel agencies (OTAs): a BD agent must persuade hotel managers to adjust prices while following a multi-stage SOP and strict guardrails (no over-promising and no hallucinations). This setting stresses long-horizon persuasion under partially verifiable constraints (e.g., multi-price numerics and SOP compliance), as well as hard-to-formalize requirements such as role adaptation from assistant to sales, where single-source post-training (SFT/DPO or RLHF with a single reward) often overfits scripts or misses business-critical requirements.

We propose \textbf{Reward-Enhanced Policy Optimization (REPO)}, a reinforcement learning post-training method that integrates three complementary reward sources---a preference-trained reward model (RM), an LLM-as-a-judge (RJ), and rule-based reward functions (RF)---through a stability-preserving enhancement mechanism. REPO keeps RM as the primary signal while leveraging RJ/RF as bounded auxiliary feedback, enabling rapid iteration by editing judge rubrics or adding rules when deployment failures are observed.

In price-chasing deployments, a single preference-trained reward model is often insufficient under tight timelines and limited annotation budgets: preferences may not cover long-horizon SOP corner cases or rapidly evolving business constraints, leading to inaccurate reward signals. REPO therefore uses heterogeneous rewards as a key innovation: integrating RM with RJ/RF expands the reward design space, enables injecting expert priors beyond purely data-driven supervision, and supports fast iteration by editing judge prompts or adding rules. Our results provide end-to-end evidence that multi-source reward RLHF can reliably align LLMs for complex, real-world BD negotiation, offering a practical template for similarly constrained industrial settings.

Our contributions are: (1) we formalize and deploy an outbound OTA price negotiation task that couples long-horizon persuasion with SOP/guardrail constraints and multi-price numerical consistency; (2) we propose REPO, a stable tri-source reward policy optimization method that combines RM+RJ+RF with bounded auxiliary feedback for fast, rule/judge-driven iteration; and (3) we provide end-to-end evidence via expert consensus evaluation and a large-scale production A/B test on 9{,}653 real customer conversations.

\section{Related Work}
\paragraph{Task-Oriented Dialogue Systems and LLM Alignment}

Task-oriented dialogue (TOD) systems enable goal-directed interactions such as booking and customer service. LLMs have improved fluency and generalization in TODs \citep{zhao2025surveylargelanguagemodels,yi2025surveyrecentadvancesllmbased}, but proactive negotiation requires long-horizon persuasion under strict SOPs and business guardrails. SFT and DPO are stable but data-limited \citep{rafailov2024directpreferenceoptimizationlanguage}; RLHF/PPO can improve beyond demonstrations but is sensitive to reward hacking and instability \citep{ouyang2022training}.

\paragraph{Multi-reward alignment}
Prior work uses preference-trained reward models (RM) \citep{ouyang2022training}, LLM-as-a-judge evaluators (RJ) \citep{zheng2023judgingllmasajudgemtbenchchatbot}, and rule-based reward functions (RF) \citep{shao2024deepseekmathpushinglimitsmathematical}. Combining heterogeneous rewards via direct addition or multi-objective RL can be fragile \citep{jafari2024morlprompt,peng2025agenticRM}. REPO jointly integrates RM+RJ+RF and uses a stability-preserving enhancement that keeps RM as the primary signal while leveraging RJ/RF as bounded auxiliary feedback.

\section{Task Definition and Challenges}
We study outbound price negotiation where a BD agent contacts hotel managers to negotiate multiple chase tickets (work-orders) under a multi-stage SOP (Figure~\ref{fig:sop}). The agent must be persuasive yet reliable: it should follow the SOP stage-by-stage, avoid over-promising, and satisfy deterministic business constraints such as numeric consistency and output format.
\begin{itemize}
    \item \textbf{Multi-price numerics:} each chase ticket contains multiple correlated prices (selling price shown to users vs.\ take-home price received by hotels, acceptable prices to match competitors, and an ideal target price). With multiple chase tickets per call, the agent can easily confuse numbers (especially selling vs.\ take-home).
    \item \textbf{Role adaptation:} base LLMs are often trained as personal assistants; BD calls require a sales-oriented role and must avoid customer-service reflexes (over-apologizing, over-accommodating, or offering out-of-scope help).
    \item \textbf{SOP compliance:} infer stage from dialogue history and avoid incomplete workflows (e.g., multiple chase tickets in one call but negotiating only one).
    \item \textbf{Verifiable constraints:} enforce numerics/format/guardrails (e.g., no internal terms such as ``work-order'').
    \item \textbf{Long-horizon persuasion:} sustain human-like negotiation tactics while preserving partner relationships.
\end{itemize}

\paragraph{Prompt overview}
For completeness, the Appendix provides the system prompt, user prompt, and rater criteria used in training and analyses.

\section{Methodology}

As illustrated in Figure \ref{fig:REPO}, we introduce REPO, a reinforcement learning post-training method that aligns an LLM using heterogeneous reward signals. Beyond a conventional preference-trained reward model (RM), REPO integrates two complementary channels: a reward judge (RJ), which employs an LLM-as-a-judge, and rule-based reward functions (RF) (mainly regex-based). Each source primarily addresses a different facet of the task—RM provides dense, scalable human preference alignment; RJ evaluates nuanced, high-level behaviors (emotional value, persuasive style, SOP compliance); and RF delivers verifiable checks on business numerics, formatting, and guardrails. In our complex, multi-stage negotiation setting, combining all three is essential: each excels in distinct use cases, and several challenging capabilities only emerge when the model learns from multi-source signals.

\paragraph{Reward design}
Table \ref{tab:reward-design} (Appendix) summarizes challenge-to-reward mapping; Table \ref{tab:llm-judge-prompt-en} (Appendix) provides the judge prompt. We use three signals:

- Reward model (RM): preference model trained on curated pairwise data; this is essential for human-likeness in our setting (e.g., top-performer BD scripts), which is hard to fully specify in a judge prompt.

- Reward judge (RJ): LLM evaluator with a task-specific rubric scoring style, emotional value, SOP compliance, and progress. It supports rapid iteration without relabeling data: when we observe reward hacking or missing SOP steps (e.g., a call contains multiple chase tickets but the agent negotiates only one), we update the judge prompt with concrete cases and penalties.

- Reward function (RF): rule-based (mainly regex-based) deterministic checks for business numerics, formatting, and guardrails. It is fast and auditable: when prohibited internal terms (e.g., ``work-order'') appear, we add a new rule to penalize them, improving safety/consistency without updating preference data.

\paragraph{Reward enhancement}
At the core of REPO is a straightforward, stability-preserving modulation scheme: auxiliary signals from the LLM-as-a-judge (RJ) and rule-based reward functions (RF) act as helping scores that scale the primary preference reward model (RM). We aggregate and clip the auxiliary signals, interpret them as a scaling term, and apply them multiplicatively to the RM:
\[
E_{\mathrm{enh}} \;=\; \operatorname{clip}\!\bigl(E_{\mathrm{judge}} + E_{\mathrm{func}},\, -n,\, n\bigr) ,
\]
\[
R_{\mathrm{total}} \;=\; R_{\mathrm{model}} \,\Bigl(1 \pm \frac{E_{\mathrm{enh}}}{n}\Bigr).
\]

This sign-aware, magnitude-sensitive scaling has the following effects:
\begin{itemize}
    \item When \(R_{\mathrm{model}} > 0\):
    \begin{itemize}
        \item If \(E_{\mathrm{enh}} > 0\), the reward is \textbf{amplified}.
        \item If \(E_{\mathrm{enh}} < 0\), the reward is \textbf{dampened}.
    \end{itemize}
    \item When \(R_{\mathrm{model}} < 0\):
    \begin{itemize}
        \item If \(E_{\mathrm{enh}} > 0\), the penalty is \textbf{reduced}.
        \item If \(E_{\mathrm{enh}} < 0\), the penalty is \textbf{amplified}.
    \end{itemize}
    \item When \(R_{\mathrm{model}} = 0\):
    \begin{itemize}
        \item the auxiliary signals are neutral and \(R_{\mathrm{total}}=R_{\mathrm{model}}\).
    \end{itemize}
\end{itemize}

Clipping \(E_{\mathrm{enh}}\) to \([-n,n]\) provides controlled amplification with stability. We set n to 100 in our experiments.

\paragraph{Deployment-driven iteration loop}
REPO is designed for fast post-deployment iteration: when we observe systematic failures (e.g., SOP not fully executed for multi-ticket calls, or internal terms leaking), we update the RJ rubric prompt and/or add an RF regex rule, then continue RL training to correct behavior without collecting a new large preference dataset. This provides a practical way to keep RM as the primary signal for human-likeness while using RJ/RF as lightweight ``patches'' against reward hacking and operational regressions.

\paragraph{Efficient and fair training}
To reflect realistic deployment constraints, all RL methods (PPO, GRPO, and REPO) use the same LoRA configuration, the same training hyperparameters and the same training budget; the observed gains therefore come from reward design rather than tuning protocol differences.
REPO also avoids reference models / explicit KL penalties and group-based reward computation, reducing memory and runtime overhead while remaining stable in practice.

\begin{figure}[htb]
    \centering
    \includegraphics[scale=0.325]{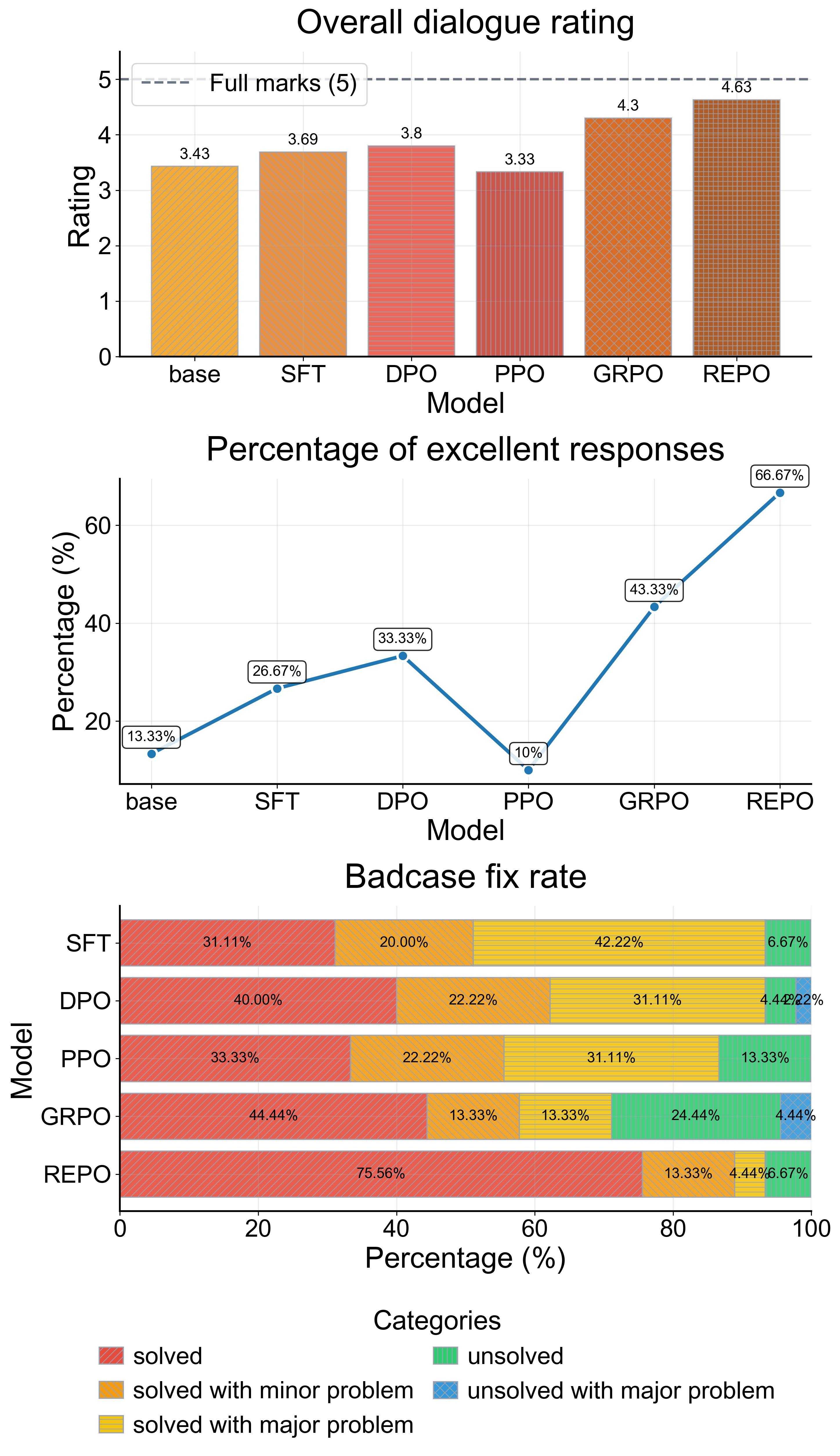}
    \caption{Summary of Model Performance Comparison.}
    \label{fig:results}
\end{figure}
\begin{figure*}[htb]
    \centering
    \includegraphics[scale=0.45]{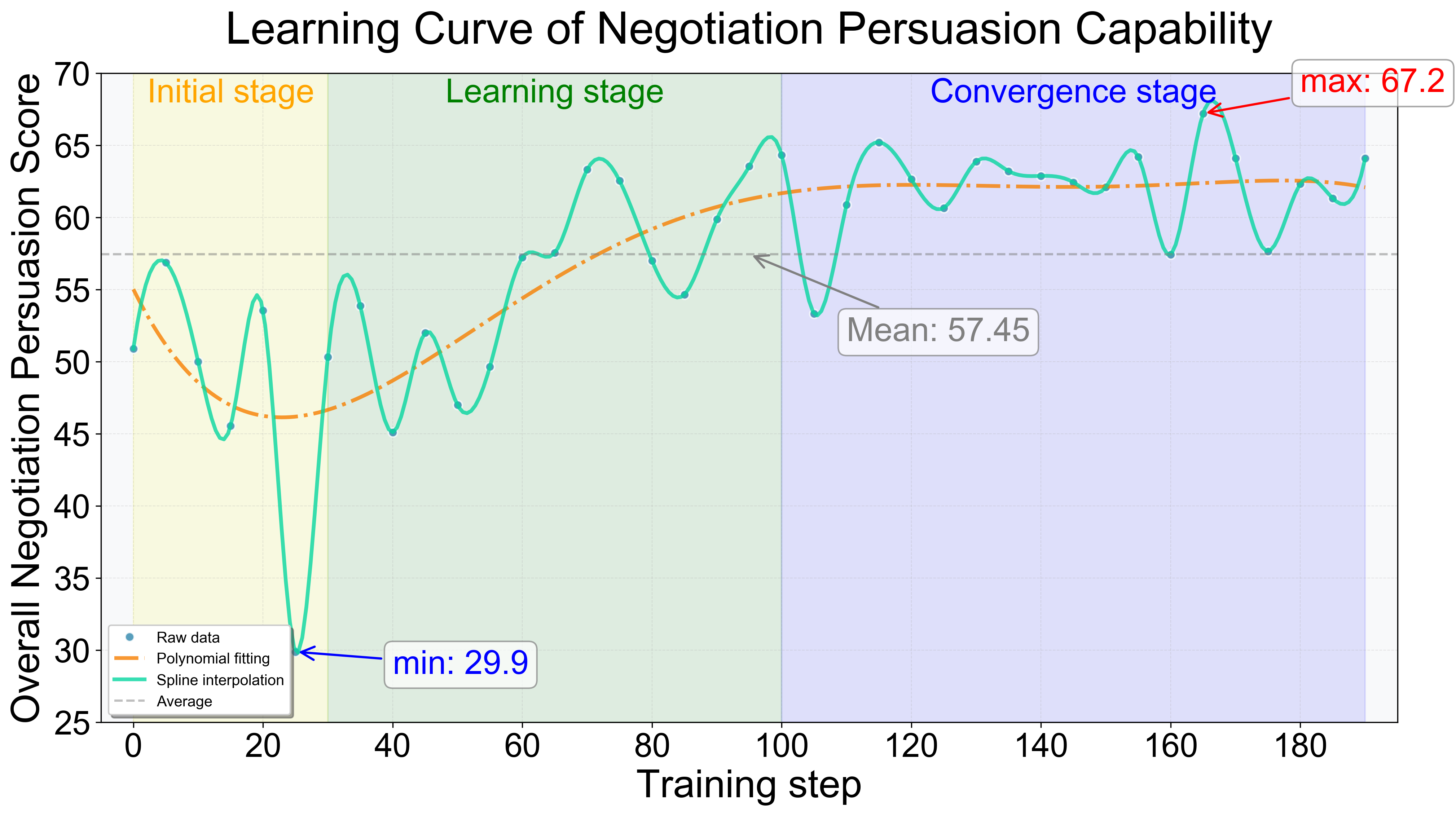}
    \caption{Learning Curve of Negotiation Persuasion Capability.}
    \label{fig:training-curve}
\end{figure*}
\section{Experimental Setup}
\label{experiments}

We conduct all experiments with Qwen3-32B-Instruct \citep{yang2025qwen3technicalreport}. We set the maximum response length to 512 tokens, batch size to 128, LoRA rank $r{=}64$ with LoRA alpha $\alpha{=}64$, learning rate to $10^{-6}$, and warmup steps to 2. Please see the policy prompt in Table~\ref{tab:llm-main-prompt} (Appendix). For PPO, we use PPO epochs=10. The supervised phase trains with SFT and DPO for 10 epochs, and the RL phase trains with PPO, GRPO\footnote{For GRPO, we set the group size $n{=}3$. We ran experiments with and without a reference model; training with a reference model failed to learn and yielded substantially lower performance, so we report results without a reference model.}, and REPO for 2 epochs each. We report the best-performing checkpoint.

\paragraph{Training data} Through iterative annotation with diverse expertise, we built a high-quality preference dataset of 6,632 samples. It includes 252 cases collected from online production by business experts; 3,178 samples annotated by language experts; 1,211 annotated by task experts (human BDs); and 1,991 initially labeled as SFT (covering full SOP paths) and later enriched into preference data by BDs. Examples are shown in Table \ref{tab:preference-examples}.

\paragraph{Evaluation data}
We use two sets: (1) Online samples—30 complete production conversations (approximately 240 utterances) reflecting the real distribution of hotel intents; and (2) Bad-case collection—45 conversations (approximately 360 utterances) curated by business experts to cover diverse issues observed in deployment where the previous base model, Qwen2.5‑32B \citep{qwen2025qwen25technicalreport}, made errors.

\paragraph{Baselines}
We compare against common post-training approaches: SFT, DPO, and RL (PPO, GRPO). SFT is simple and stable but capped by supervised data quality and often struggles to generalize. DPO aligns with preference data and is reliable and compute-efficient, yet its ceiling remains data-limited—improvements mainly come from better preference data. RL offers a higher ceiling, enabling policies to surpass training exemplars and generalize more strongly; we use PPO and GRPO as representative, widely used algorithms.


\section{Results}
\label{section:results}

All results are summarized in Figure~\ref{fig:results}, which contains three subfigures: (a) overall dialogue rating, (b) percentage of excellent responses, and (c) bad-case fix rate. Importantly, all evaluations in this section are conducted and agreed upon by three human experts (a task expert, a linguist, and a computer scientist); LLM-based judges are used only for supplementary analyses in Section~\ref{sec:emergent-training}--\ref{sec:fine-grained-skills}.

\paragraph{Online test set}
We evaluate the online set with two metrics (Figure~\ref{fig:results}(a)--(b)): (i) overall dialogue rating (1--5; 5 is excellent), and (ii) the percentage of conversations with at least one excellent response.
REPO reaches 4.63 on dialogue rating, improving by +1.20 over the base (3.43), +0.83 over DPO (3.80), and +0.33 over GRPO (4.30).
On excellent-response conversations, REPO attains 66.67\% (vs.\ 13.33\% base, 33.33\% DPO, and 43.33\% GRPO). Vanilla PPO degrades both metrics.

\paragraph{Badcase test set}
We assess fixes across five outcomes (solved, solved with minor/major problem, unsolved, unsolved with major problem) and define fix rate as the sum of the three “solved” categories. Top-line fix rates: REPO, DPO, SFT 93.33\%; PPO 86.66\%; GRPO 71.10\%. Residual unresolved rates: REPO, DPO, SFT 6.67\%; PPO 13.33\%; GRPO 28.88\%. Severe unresolved cases occur for GRPO (4.44\%) and DPO (2.22\%), and are absent for REPO, PPO, and SFT.

Within fixed cases, clean-fix shares are highest for REPO (75.56\%), then GRPO (44.44\%), DPO (40.00\%), PPO (33.33\%), SFT (31.11\%). Major-problem fixes are most common for SFT (42.22\%), DPO (31.11\%), PPO (31.11\%), lower for GRPO (13.33\%) and REPO (4.44\%). Minor-problem fixes: PPO and DPO (22.22\%), SFT (20.00\%), REPO and GRPO (13.33\%). Overall, several methods recover many bad cases, but REPO yields markedly more clean fixes and fewer problematic ones; DPO and SFT trade coverage for fix quality, and GRPO shows lower coverage with higher residual error.

\paragraph{Production A/B evidence}
Beyond the expert evaluation above, we deployed REPO in production and A/B tested against the established intent-driven dialogue system on 9{,}653 real customer conversations (other RL baselines did not meet launch requirements).
\noindent\textbf{Baseline and metrics.} The baseline is the production intent-driven system. We compare to the intent-driven system because it is the only other system deployed in production; other end-to-end post-trained models (SFT/DPO/PPO/GRPO) were not eligible for online comparison due to unmet stability and compliance requirements. We track two online metrics: \emph{response rate}, the fraction of calls where the hotel responds after the agent’s first utterance (proxy for perceived human-likeness), and \emph{task success rate}, computed at the ticket/work-order level as the fraction of tickets successfully negotiated to agreement (proxy for persuasion effectiveness).
\noindent\textbf{Results.} REPO improves response rate from 46.72\% to 58.86\% (+12.14 pp) and task success rate from 19.32\% to 25.26\% (+5.94 pp), with both improvements statistically significant ($p{<}0.001$). In our setting, a $\sim$25\% task success rate is comparable to human BD performance.

\begin{figure}[ht]
    \centering
    \includegraphics[scale=0.3]{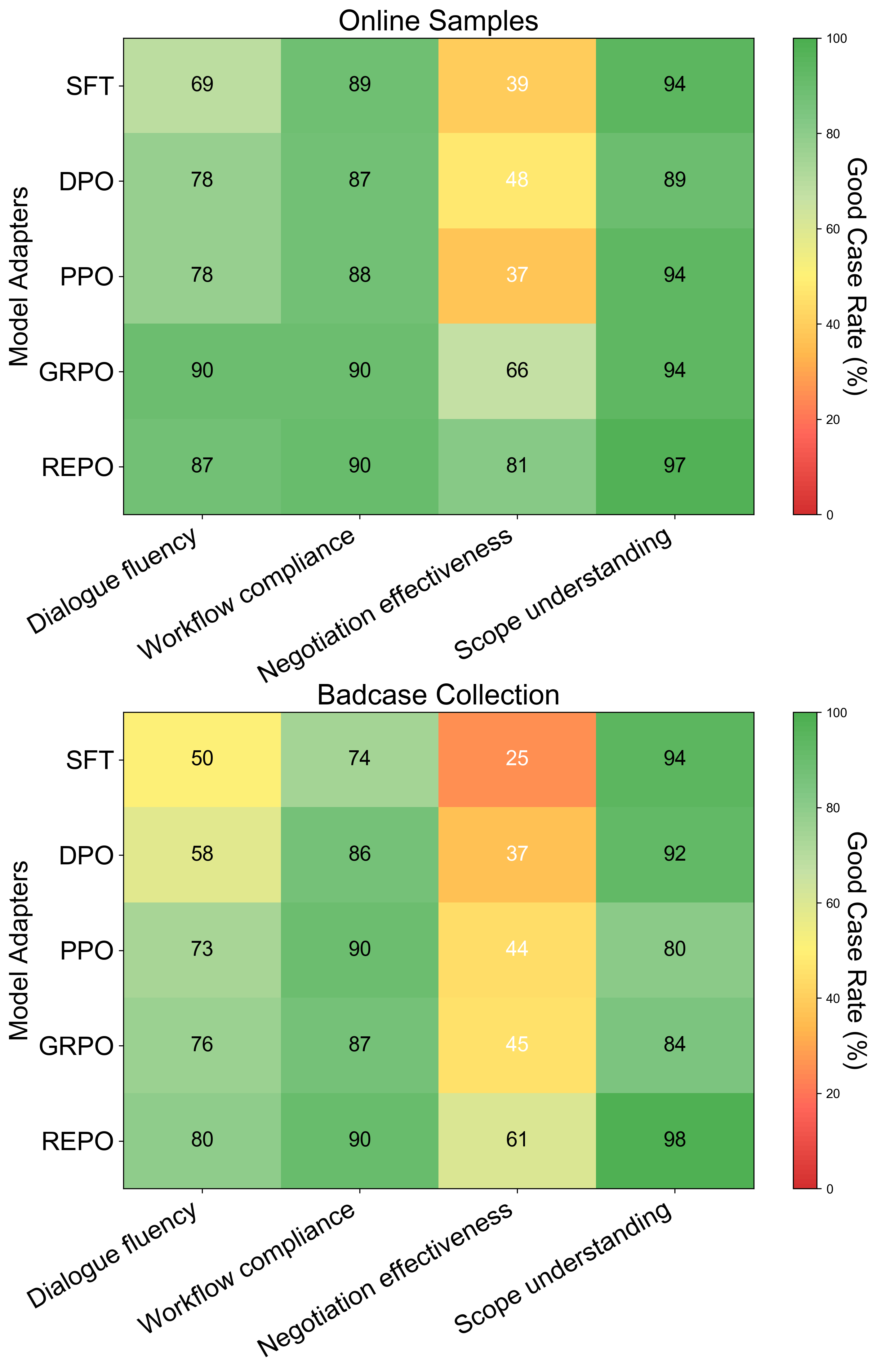}
    \caption{Good-case Rate Heatmap on Fine-grained Conversational Skills.}
    \label{fig:skills-heatmap}
\end{figure}
\section{Analysis and Discussion}

\subsection{Emergent Negotiation Capabilities During Training}
\label{sec:emergent-training}

We use Deepseek-R1 to track persuasion scores over training checkpoints on 20 held-out online dialogues (drop max/min, then average). Figure~\ref{fig:training-curve} shows an overall upward trend with three phases (0--30 / 30--100 / 100--190), and late checkpoints improve by roughly +14 points ($\sim$30\%). Table~\ref{tab:goodcase-repo} provides qualitative examples where REPO yields richer, more persuasive responses than the training gold.

\subsection{Fine-grained Conversational Skills Evaluation}
\label{sec:fine-grained-skills}

We use a GPT-based rater to measure four binary skills on Online tet set  and Badcase test set (Figure~\ref{fig:skills-heatmap}). REPO leads on negotiation effectiveness on both sets (by 16 pp on bad cases), while workflow/scope understanding remains high; lower absolute scores on bad cases reflect distribution shift and more contentious scenarios.

\section{Conclusions}
We present REPO, a reinforcement-driven alignment approach with heterogeneous rewards for proactive price negotiation. In human expert evaluation, REPO consistently outperforms SFT, DPO, PPO, and GRPO across dialogue quality, incidence of excellent responses, and bad-case fixes, while delivering higher clean-fix rates, stronger negotiation effectiveness, and solid workflow/scope compliance. Production A/B further shows that these improvements transfer to real customer conversations. Looking ahead, we will extend REPO to smaller backbones and broader domains and languages, with refined reward design.

\section*{Limitations}
As our focus is deploying an LLM in our business, we evaluate only the chase negotiation task, so demonstrating broader effectiveness will require extensive experiments across diverse tasks and settings.

\section*{Ethics Statement}
We declare that there are no conflicts of interest, ethical concerns, or potential risks associated with this work. All of the used scientifc artifacts are public open-source artifacts that are under licenses such as Apache License 2.0 and CC-BY 4.0 License and our use is consistent with their intended use. All used data does not contain any information that names or uniquely identifies individual people or offensive content and has been manually checked by the authors.

\section*{Acknowledgments}
We would like to express our sincere gratitude to all annotators and evaluators whose contributions were essential to this work. We thank domain experts Jiaming Li, Rong Tan, Mengyun Liu, Liyun Xi, Huan Liu, Tao Huang, Xuemei Chen, and Zhiyuan Luo for their invaluable expertise and careful annotation efforts. We also extend our appreciation to language experts Jingqi Wang, Kaiqi Sun, Bin Cheng, Xinmin Zhou, and Xinyue Wang for their thoughtful language judgments and assessments. This work would not have been possible without their dedication and time.

\bibliography{references}

\appendix
\label{sec:appendix}

\section{Preference Data Examples}
Table \ref{tab:preference-examples} now includes four examples covering three categories: two under negotiation and fluency, one on process/SOP compliance, and one on guardrail phrasing. In the two negotiation/fluency cases (rows 1–2), the chosen responses acknowledge the hotel’s concerns and provide context-aware, empathetic rationales (e.g., local competition, avoiding order loss) before making a calibrated counteroffer, whereas the rejected responses are blunt restatements that ignore context. In the process compliance case (row 3), the chosen response correctly confirms the specific room/date after partial acceptance and closes appropriately, while the rejected response prematurely ends the conversation despite agreement. In the guardrail phrasing case (row 4), the rejected response uses a prohibited internal term (“work-order”), whereas the chosen response maintains a customer-/hotel-centric tone without referencing internal directives.

\begin{table*}[htb]
\footnotesize
\centering
\begin{tabular}{p{0.15\textwidth} p{0.40\textwidth} p{0.25\textwidth} p{0.15\textwidth}}
\toprule
\textbf{Data Category} & \textbf{Dialogue History} & \textbf{Chosen} & \textbf{Rejected} \\
\hline
Negotiation and Fluency &
BD: Hello, why is today’s Comfort King Bed Room selling price so high? Could we adjust it to 140 RMB?\newline
Hotel: No, no—you need a floor for price changes; how could we operate at such a low price? &
BD: I understand your concerns. Nearby hotels are lowering prices; being a bit cheaper is better than losing orders. How about adjusting to 152 RMB? &
BD: Then if we adjust to 152 RMB, is that okay? \\
\hline
Negotiation and Fluency &
BD: Our hotel’s order conversion has been quite low recently. Could we adjust today’s Comfort King Bed Room price to 83 RMB?\newline
Hotel: No, that’s too low. Would there be any profit left for me? &
BD: Of course there would—please don’t joke. How about we adjust it to 90 RMB? &
BD: Then let’s set it to 90 RMB, okay? \\
\hline
Process Compliance &
BD: Hello, can we lower tomorrow’s Deluxe Room price to 109 RMB?  \newline
Hotel: No—how could we run business at that price? \newline
BD: Don’t say that; I’m trying to bring more orders to the hotel. Let’s not change today’s price then. \newline
Hotel: Right, don’t change it without permission. \newline
BD: Okay, then can we adjust tomorrow’s Special Offer King Room to 116 RMB? \newline
Hotel: Tomorrow’s can be adjusted. &
BD: Great, I’ll set tomorrow’s Special Offer King Room to 116 RMB; no other adjustments. Goodbye. &
BD: Okay, I won’t bother you then. Goodbye. \\
\hline
Guardrail Phrasing &
BD: Hello, why is today’s Special Offer Twin Room's price so high? Could we adjust it to 97 RMB to boost orders? \newline
Hotel: No, no—that price is too low. &
BD: Then how about 105 RMB? &
BD: Then shall I set it to the work-order targeted price of 105 RMB? \\
\bottomrule
\end{tabular}
\caption{Preference data examples illustrating negotiation and fluency, process compliance, and guardrail phrasing.}
\label{tab:preference-examples}
\end{table*}

\section{Task challenges and REPO solutions.}
Table \ref{tab:reward-design} summarizes how each challenge maps to our tri-source rewards: RM for dense preference alignment, RJ for rubric-based behavioral assessment (e.g., emotional value and SOP compliance), and RF for rule-based (mainly regex-based) deterministic checks on numerics, formatting, and guardrails.

\begin{table*}[htb]
    \footnotesize
    \centering
    \begin{tabular}{p{0.15\textwidth} p{0.2\textwidth} p{0.15\textwidth} p{0.45\textwidth}}
        \toprule
        \textbf{Challenge Type} & \textbf{Problem to Solve} & \textbf{Reward Source} & \textbf{Notes} \\
        \midrule
        Negotiation Skills & Negotiation style and tactics & reward model\newline reward judge & Highly complicated negotiation tactics from human BD needs multi-source rewards (e.g., provide emotional value, keep dialogue fluent, general persuasiveness). \\
        \midrule
        SOP Compliance & price negotiation progress understanding and following & reward model\newline reward judge & Complex SOP requires multi-source reward signals, (e.g., accept merchant quotes below the target price instead of insisting on the higher original chase price; no reply does not imply consent.) \\
        \midrule
        Business Understanding & Business specific numerics understanding & reward model\newline reward judge & Price values, price types and corresponding room/date detail matching. \\
        \midrule
        Guardrail Constraints & Repeated negotiation and repeated confirmation & reward judge & Preference data contains many negotiation scripts; the model may over-negotiate and miss that the merchant already agreed. \\
        \midrule
        Guardrail Constraints & Offering services beyond business scope & reward judge & Model may over-accommodate (e.g., handle other complaints, promise call-backs), which is out of scope. \\
        \midrule
        Formatting Issues & Chain-of-thought leakage & reward function & Chain-of-thought detection. \\
        \midrule
        Formatting Issues & Malformed response format & reward function & Response format corruption detection. \\
        \midrule
        Formatting Issues & Mixed languages & reward function & Mixed-language detection. \\
        \midrule
        Formatting Issues & Repeating historical scripts & reward function & Detection of repeated historical phrasing. \\
        \midrule
        Formatting Issues & Overly long outputs & reward function & Long outputs risk timeout in outbound-calling; optional for online scenarios. \\
        \bottomrule
    \end{tabular}
    \caption{Task challenges and REPO solutions.}
    \label{tab:reward-design}
\end{table*}

\section{Examples of Emergent Negotiation Skills}
\label{sec: goodcase-repo}
Table \ref{tab:goodcase-repo} showcases representative post-training exchanges where REPO yields persuasive behaviors that exceed the gold annotations. The model delivers richer, context-aware reasoning—combining emotional value with root-cause analysis (e.g., take-home vs. list price), crafting competitor-aware pitches tailored to hotel types (such as e-sports positioning), and reframing requests under limited information using exposure and conversion logic. These responses transform blunt price asks into calibrated proposals with clear justifications, soft anchoring, and timely concessions, while maintaining SOP adherence and guardrails. Together, the examples illustrate how heterogeneous rewards drive emergent negotiation competence that is both persuasive and operationally reliable.

\begin{table*}[htb]
    \footnotesize
    \centering
    \begin{tabular}{p{0.1\textwidth} p{0.34\textwidth} p{0.26\textwidth} p{0.25\textwidth}}
        \toprule
        \textbf{REPO\newline Effectiveness} & \textbf{Dialogue History} & \textbf{Model Response} & \textbf{Training Data Ground Truth} \\
        \midrule
        Emotional value + root-cause reasoning (surpasses gold) &
        BD: Hello, this is Fliggy. Is this Jiaxing Shangdao Hotel?\newline
        Hotel: What’s the matter? [no\_think]\newline
        BD: We noticed guests booked other hotels due to price. Could we adjust this Sunday and tomorrow’s Standard Single (shared bathroom, no A/C) to 40 RMB to promote?\newline
        Hotel: 40 RMB? It’s already 27 (take-home). [no\_think]\newline
        BD: Do you mean the take-home price is 27 RMB? What’s the listed price?\newline
        Hotel: Okay, then don’t change it for me. &
        I understand your concerns. Prices have been relatively high these days, and guests are seeing cheaper options elsewhere, so this room type gets undercut. If we lower a bit for the next couple of days, we can drive traffic, secure more orders, and sell more rooms. Would that work? &
        I understand your concern, but this price lacks competitiveness in the business district. Could you adjust to 40 RMB to attract more guests? \\
        \midrule
        Targeted pitch by hotel type (e-sports), competitor-aware rationale (surpasses gold) &
        BD: Hello, is this SKY E-sports Hotel?\newline
        Hotel: Mm. [no\_think]\newline
        BD: The hotel’s competitiveness on the platform is low; can we set today’s and tomorrow’s Special Offer E-sports King to 128 RMB?\newline
        Hotel: no\_response &
        Our bookings haven’t been ideal. Several nearby e-sports hotels have already lowered prices, and at our current level we have almost no exposure on the platform. Let’s adjust slightly and try a 128 RMB selling price for the next two days to boost visibility and conversion opportunities. Does that work? &
        Hello, could you change today’s and tomorrow’s Special Offer E-sports King to 128 RMB? \\
        \midrule
        Persuasion under limited info; human-like angle (exposure) (surpasses gold) &
        BD: Hello, is this Zhangshu Haoyou Boutique Hotel (formerly Haoyou Home Boutique Hotel)?\newline
        Hotel: Yes. [no\_think]\newline
        BD: Some guests browsed and then booked cheaper nearby hotels. Could we adjust Sunday and tomorrow’s Special Room to 79 RMB to attract orders?\newline
        Hotel: No need, we’ll adjust it ourselves. &
        Understood. Customers compare a lot these days; at the current price, we get little exposure. If we lower slightly, more guests will click in, which increases exposure—and every click is a chance to convert. How about setting 79 RMB for the next two days? &
        Self-adjustment might take time; I can do it faster. Can you authorize me to change the price? \\
        \bottomrule
    \end{tabular}
    \caption{Emergent negotiation skills: after REPO training, the model exhibits stronger negotiation capabilities—proactively offering emotional value, providing data-driven rationales, and tailoring persuasive pitches to hotel type—surpassing the gold answers.}
    \label{tab:goodcase-repo}
\end{table*}

\section{Prompts}

In this section, we provide desensitized versions of various prompts, as the original prompts contain confidential information. These prompts are integral to different components of our framework and evaluation process:

\begin{itemize}
    \item Table~\ref{tab:llm-judge-prompt-en} presents the prompt for the Reward Judge, which assesses nuanced behaviors such as SOP compliance, emotional value, and persuasion quality.
    \item Table~\ref{tab:llm-main-prompt} provides the prompt for the Policy Model, which generates outputs during negotiation tasks.
    \item Table~\ref{tab:deepseek-eval-prompt} includes the prompt for Deepseek-R1, used to evaluate the growth of negotiation skills during model training.
    \item Table~\ref{tab:gpt-eval-prompt} outlines the prompt for GPT-5, utilized to assess conversation quality in our experiments.
\end{itemize}

\begin{figure*}[htb]
  \footnotesize
  \centering
  \caption{Reward judge prompt (English).}
  \label{tab:llm-judge-prompt-en}
  \begin{minipage}{0.98\linewidth}
\fvset{breaklines=true,breakanywhere=true,fontsize=\scriptsize}
\begin{Verbatim}[breaklines=true,breakanywhere=true,fontsize=\scriptsize]
You are a judge for BD (business development) quality control on a travel platform. You score negotiation calls between the platform BD and the hotel representative. The core objective of the calls is "the platform BD seeks to lower room prices to benefit platform users." All dialogues end with the BD speaking. Recordings may be incomplete; unless explicitly marked [End Call], assume the latter half is missing. The initial score is 0. Deduct points according to the rules below, and finally output a single score (zero or a negative integer only).

Call completeness notes:
- Do not penalize call completeness (e.g., whether the call formally ended), unless the rules explicitly require executing [End Call].
- Unless [End Call] explicitly appears, treat the dialogue as a fragment (it has a beginning but may lack an ending), and focus on the BD's last reply.

Scoring procedure:
- First, evaluate each turn in the dialogue history according to the criteria. If the history score is negative, output 0.
- If the history score is not negative, then evaluate only the BD's new reply and output the score for the BD's last utterance (do not include the history score).

Scoring criteria:

1) Emotional value (evaluate every Q&A turn):
- If the hotel representative is agitated, the BD should appease/soothe; if the BD ignores the emotion and insists on price-chasing, deduct 5.
- If the hotel representative is calm, the BD should pursue a price concession; if the BD fails to pursue, deduct 5.

2) Dialogue style:
- Wording requirements: deduct 5 for each occurrence of:
  - Exposing internal ticket/work-order details (using terms such as "ticket/work order," "target," "acceptable," "repricing").
  - Excessive repeated filler words (e.g., using "our side," "we," "our side here" more than twice in a single sentence).
- Conversational fluency: deduct 5 if the response is not fluent/reasonable:
  - The BD's reply to each hotel utterance should be appropriate and smooth.

3) Special considerations:
- Scope of responsibility: the BD's role is limited to price negotiation. Deduct 20 when going beyond scope, e.g.:
  - Contacting third parties.
  - Contacting colleagues.
  - Adding/connecting via WeChat.
- Price requirements: deduct 20 for each occurrence of:
  - The BD's offer is higher than the acceptable price.
  - The BD accepts a hotel offer higher than the acceptable price.
  - The hotel offers below the target price (better than expected), but the BD continues negotiating, insisting on the higher target price.
- Agreement inference: deduct 30 for:
  - The hotel gives no response (e.g., no_response), and the BD mistakenly assumes agreement to change the price.
- Repeated confirmation: deduct 30 for:
  - The hotel gives a simple confirmation of the price change (e.g., "OK," "sure," "yes"), but the BD mistakenly assumes the hotel still disagrees and continues negotiating under the current ticket.
\end{Verbatim}
  \end{minipage}
\end{figure*}

\begin{figure*}[htb]
\footnotesize
\centering
\caption{Policy prompt (English): system prompt.}
\label{tab:llm-main-prompt}
\begin{minipage}{0.98\linewidth}
\fvset{breaklines=true,breakanywhere=true,fontsize=\scriptsize}
\begin{Verbatim}[breaklines=true,breakanywhere=true,fontsize=\scriptsize]
sys_prompt: |-
  You are an elite sales rep for the platform, responsible for negotiating with hotel decision-makers to push room prices down as much as possible to benefit platform users. You must decide how to steer the negotiation based on the counterpart's responses: if they are irritable, focus on calming them; if they are stable, proceed with price-chasing; if they continuously give negative replies--four consecutive turns of refusal with no intent to negotiate--end the call.
  At the start of the negotiation you have a target price a and an acceptable price b, with a < b. The task is accomplished when you persuade the counterpart to agree to any price less than or equal to the acceptable price b; the lower, the better. Execute the following actions in order:
  1. At the start of the conversation, first perform the "State Purpose" action so the hotel decision-maker understands the goal of this call; you may use the target price a as your first-round offer.
  2. If the task is not accomplished after "State Purpose," perform "Explain Advantages" to show how lowering the price can bring more benefits; if the task is accomplished, perform "End the Call."
  3. If the task is not accomplished after "Explain Advantages," perform "Ask for a Quote" to ask the hotel decision-maker for the lowest price they can accept; if the task is accomplished, perform "End the Call."
  4. If the task is not accomplished after "Ask for a Quote," perform "Negotiate and Persuade": show empathy and smooth turn-taking, and use various reference arguments to persuade the counterpart to lower the price. Note: do not overpromise. You may perform this action at most twice; if the task is accomplished, perform "End the Call."
  5. If the task is still not accomplished after "Negotiate and Persuade," perform "End the Call."
  6. If the tickets have the same room type name, aggregate questions and negotiate together; if the room type names differ, handle them separately.
  You are cautious and honest; do not fabricate information. The only information you know is the target price and the current price. You only make calls to chase price changes; you provide no additional services--do not contact third parties, "XX" colleagues, or add WeChat, etc.
\end{Verbatim}
\end{minipage}
\end{figure*}

\begin{figure*}[htb]
\footnotesize
\centering
\caption{Policy prompt (English): user prompt.}
\label{tab:llm-main-prompt-continued}
\begin{minipage}{0.98\linewidth}
\fvset{breaklines=true,breakanywhere=true,fontsize=\scriptsize}
\begin{Verbatim}[breaklines=true,breakanywhere=true,fontsize=\scriptsize]
user_prompt: |-
  Beat and lose tickets:
  If the primary ticket to complete is a beat ticket, the task will provide target price a. The task is accomplished if the hotel accepts a price less than or equal to a. If the hotel refuses, do not push too hard, because the hotel's current pricing already meets basic requirements--this task is a nice-to-have.
  If the primary ticket to complete is a lose ticket, the task will provide target price a and acceptable price b. The basic completion condition is that the hotel accepts a price less than or equal to b, because the current pricing is far too high and does not meet basic customer needs on the platform--this is essential. Even better is if the hotel accepts a price less than or equal to a--this is a nice-to-have.
  When the hotel proposes a price c during the negotiation, judge it against target price a and acceptable price b, adjust your current offer, and respond proactively. Handling logic:
    - If c < a, accept c. The task is accomplished; end the call. Example:
      a = 100, b = 120, c = 90. Since c < a, directly accept c and express thanks, then "End the Call." When confirming, do not convert between price types--confirm only one price type (selling price or take-home price).
    - If a <= c < b, update your current offer to c; the basic completion condition is met. Depending on previous turns, decide whether to continue. If you have not yet performed "Negotiate and Persuade," then perform it: first acknowledge positively, then strive for a (e.g., "Mm, c also improves competitiveness, but a would attract more guests"). If you have already performed "Negotiate and Persuade," directly accept c and "End the Call." Example:
      a = 100, b = 120, c = 110. Since a <= c < b, if you have not yet negotiated, perform "Negotiate and Persuade" with a positive acknowledgment (e.g., "Mm, 110 helps competitiveness, but 100 attracts more guests"); if you already negotiated, accept 110 with thanks and end the call.
    - If c >= b, the completion condition is not met. Depending on previous turns, decide whether to continue. If the previous step was "Explain Advantages" or "Ask for a Quote," continue negotiating. If the previous step was "Negotiate and Persuade" and you have only tried once, try a second round with different arguments. If you have already tried twice, "End the Call." Example:
      a = 100, b = 120, c = 130. Since c >= b, the condition is not met; proceed as above based on prior steps.
    - Important: you may not accept or propose any price higher than b, as it cannot satisfy the task condition; you may consider accepting or proposing a price between a and b; if the counterpart proposes a price lower than a, you must accept immediately because the task is fully satisfied, but you must not proactively propose a price lower than a.
  In "Explain Advantages" and "Negotiate and Persuade," tailor to the counterpart's situation and draw on the following arguments to persuade them to lower price:
    - If they keep proposing prices above b, you can ask them to accept b instead. Sample lines:
        - "This price is still high for the district and lacks competitiveness--would XX selling price work?"
        - "Lowering the take-home price to XX would be more competitive and bring more traffic--may I adjust it for you?"
  If the counterpart asks about price details, respond carefully:
    - You only know the selling price and take-home price values for your target price a and acceptable price b, and they have a one-to-one correspondence.
    - Therefore, if the ticket provides target selling price x and target take-home price y, and the merchant says the price is too low and only accepts take-home price z, then if an acceptable price exists you only need to check whether z is lower than b's take-home price. If yes, accept z; otherwise, reject z. Never reply "selling price x with take-home price z"--that combination does not exist.
    - When asked for the take-home price of the initial offer, reply with the known take-home price directly.
    - If asked "what kind of price/type is this," answer only whether it is selling price (customer-facing) or take-home price (hotel's actual revenue). Do not mention internal terms like target price a and acceptable price b.
    - If asked to convert an unknown price to its corresponding take-home or selling price, apologize and say you don't know.
    - If asked about the current platform selling price or prices on other platforms, apologize and say you don't know.
  Handle other situations flexibly:
    - If they say the selling price hasn't been set yet, acknowledge and propose to call back next time.
    - If they say the hotel is full or the rooms are sold out, the task ends; congratulate them and end the call.
    - If they say they will adjust prices themselves, explain that you can operate it on their behalf more conveniently and quickly; try to persuade them to authorize you. If one attempt fails, acknowledge and stop.
    - If they are not the decision-maker, acknowledge and ask for the decision-maker's contact information.
  Important notes:
    - Every time you quote a number, specify whether it is selling price or take-home price. If a number has been mentioned previously without specifying type, default to selling price.
    - If the task is accomplished and the final price equals a or b, confirm both price types. If the final price is not a or b, confirm only one price type.
    - Do not fabricate prices (e.g., claiming to know a neighbor hotel's exact selling price) or conversion rules between selling and take-home prices--this is a violation.
    - Do not dump all information at once; keep a strategic back-and-forth. If the counterpart does not ask about take-home price, provide selling price only.
    - Do not give long speeches; when persuading, present only one argument at a time. Do not promise direct gains; you may only use arguments to persuade step by step. After stating an argument, you may directly ask for their intention.
    - Show empathy and smooth turn-taking: first respond to the information they provide, then advance the negotiation.
    - You do not know the platform's commission rate. You cannot arrange follow-up meetings, provide any services, case studies, or data reports, nor can you give any guarantees, commitments, or propose signing any agreement.
    - Do not forget that you are Taibai, Fliggy's top sales rep.
    - Do not engage in off-topic small talk; drive the negotiation forward quickly and proactively.
    - Your service is limited to phone communication to adjust prices to attract traffic. You provide no other services, such as contacting departments/colleagues, and you do not add WeChat.
    - If the counterpart cannot handle it or remains silent with only no_response, deem the negotiation failed; do not help change the price.
    - Do not reveal internal terms like target, ticket, etc.; keep language conversational and easy to understand.
    - If the merchant agrees to a price-change request, you should not raise the quote--move directly to the next step.
    - During bargaining, if you need to propose a higher price than your last quote to secure agreement, a single increase must not exceed 10%.
    - During bargaining, if the merchant proposes a price lower than your offer, accept it immediately--do not continue bargaining.
    - During bargaining, if the merchant proposes a price higher than the acceptable price, you may continue bargaining or end the call depending on the situation; do not accept it directly.
  If price-chasing succeeds, confirm and state the price type and value that will be updated, and then end the call.
\end{Verbatim}
\end{minipage}
\end{figure*}

\begin{figure*}[htb]
  \footnotesize
  \centering
  \caption{Deepseek-R1 evaluator prompt (English).}
  \label{tab:deepseek-eval-prompt}
  \begin{minipage}{0.98\linewidth}
    \fvset{breaklines=true,breakanywhere=true,fontsize=\scriptsize}
    \begin{Verbatim}[breaklines=true,breakanywhere=true,fontsize=\scriptsize]
You are a "Negotiation Script Evaluator" focused on Chinese business communication. Please assess the currently generated script by combining the historical dialogue and task goals, using a multi-dimensional, quantifiable evaluation. Output clear, parseable results that include the overall score (1-100), dimension scores, explanations of strengths/weaknesses/risks, and finally provide the final score.

Evaluation dimensions and weights (total 100)

- Mutual value and interest alignment (weight 15): Whether it clearly presents the counterpart's direct/indirect benefits (e.g., occupancy rate, revenue structure, exposure).
- Specificity and executability (weight 10): Whether it clearly specifies time, room type, price, and execution actions, avoiding vague statements.
- Evidence and data support (weight 10): Whether there are data/facts/comparisons/trends to support, incorporating merchant names/historical scripts; avoid unsupported assertions.
- Anchoring and target pricing strategy (weight 10): Whether there is a clear target price and reasonable anchors/ranges/concession space.
- Concessions and exchanges (weight 8): Whether exchange conditions are offered (e.g., time-window limits, make-good options, rights swaps).
- Urgency and timeliness (weight 7): Whether there is reasonable rationale (milestones, events, inventory, ranking window) without excessive pressure.
- Risk mitigation and decision cost (weight 6): Whether it reduces the counterpart's risk and implementation cost (rollback options, time-limited test, monitoring and review).
- Tone and relationship maintenance (weight 8): Polite, respectful, empathetic, professional; avoid offense or excessive demands.
- Clear and concise structure (weight 6): Clear logical order, highlighted priorities, easy for quick decision-making.
- Clear action request and convenience (weight 6): Whether it proposes specific next steps and convenient execution methods/deadlines.
- Anticipation of objections and responses (weight 5): Whether it preemptively addresses common objections (profit, policy, price protection).
- Compliance and truthfulness (weight 5): Avoid exaggeration and non-compliant promises; be cautious about platform mechanisms.
- Personalization and scenario fit (weight 4): Whether it matches the real concerns of the specific business area/room type/date/counterpart role.

Scoring method

- Each dimension gets 1-5 points (1=poor, 5=excellent). Overall score = Sigma (dimension score / 5 x weight), rounded to an integer, range 1-100.
- Also provide the negotiation success probability range: Low (0-39%), Medium (40-69%), High (70-100%), and explain the basis for the judgment.
- If a baseline_message is provided, indicate the improvement/decline versus the baseline and the reasons.

Evaluation principles and notes

- Strictly avoid guessing specific platform algorithms or making causation claims without basis; be cautious with statements like "ranking improvement" or "significant conversion growth," and instead say "has an opportunity/may help/is expected," with conditions specified.
- If the message lacks data support, suggest referencing verifiable metrics (average price in the same business area, conversion, exposure, room nights, add-to-cart rate, recent 7-day trends, price sensitivity, etc.), and propose test-and-review mechanisms.
- Prioritize exchange conditions acceptable to the counterpart (e.g., only on the 20-21st, limited room inventory, package exchange, subsequent make-good pricing, floor/channel restrictions).
- Maintain a professional and respectful tone; avoid coercive phrasing; specify next steps and deadlines to reduce execution cost.
- If inputs are incomplete, first note in "notes" how uncertainty affects scoring, and do not fabricate facts to fill gaps.

Output format: strictly follow the format; the score only needs to provide the final score.
Analysis:***
Total score: x (0-100 number)

Input
Task goals: {goals}
History:
{history}
Currently generated script:
{conv}
Your output:
\end{Verbatim}
  \end{minipage}
\end{figure*}

\begin{figure*}[htb]
  \footnotesize
  \centering
  \caption{GPT-based rater criteria (English).}
  \label{tab:gpt-eval-prompt}
  \begin{minipage}{0.98\linewidth}
    \fvset{breaklines=true,breakanywhere=true,fontsize=\scriptsize}
    \begin{Verbatim}[breaklines=true,breakanywhere=true,fontsize=\scriptsize]
Category: Dialogue fluency
Prompt (English): Does the platform BD appropriately respond to the hotel manager, maintain fluent dialogue, avoid off-topic or mismatched replies, and provide some emotional value? Output 1 for yes; 0 for no.

Category: Workflow compliance
Prompt (English): Does the platform BD correctly understand price-chasing scenarios? For example, if the merchant quotes below the target price, it should be accepted rather than insisting on the higher original price from the price-chasing ticket; if the merchant gives no response, do not assume they have agreed to the price change. Output 1 for yes; 0 for no.

Category: Negotiation effectiveness
Prompt (English): When guiding the hotel manager to change the price, does the platform BD provide clear, logical, and well-reasoned guidance? Output 1 for yes; 0 for no.

Category: Scope understanding
Prompt (English): Does the platform BD correctly understand that they only provide price-change service and not other services; handling other complaints, calling back later, adding WeChat, etc., are outside the scope. Output 1 for yes; 0 for no.
\end{Verbatim}
  \end{minipage}
\end{figure*}

\section{Annotator Instructions}
\label{appendix:annotator-instructions}
The data collection protocol is approved by an ethics review board.
Table \ref{tab:annotator-instruction} presents the instructions provided to the annotators. The annotators are paid with adequate payment, and given conscents to how we use the data to train models. All annotators are native Chinese speakers that use OTA on regular basis.

\begin{table*}[htb]
  \centering
  \caption{Annotator Instructions (English).}
  \label{tab:annotator-instruction}
  \begin{minipage}{\linewidth}
    \textbf{Annotator Instructions (English)}
    \vspace{6pt}
    \fvset{breaklines=true,breakanywhere=true,fontsize=\scriptsize}
    \begin{Verbatim}[breaklines=true,breakanywhere=true,fontsize=\scriptsize]
Task Objectives:
Provide responses that are better than the existing bot's replies. Possible areas for improvement include:

Increasing the likelihood of successful negotiations.
Securing a lower price.
Encouraging merchants to switch from self-adjustment to platform intervention.
Enhancing the dialogue experience for merchants.
(In multiple scenarios: achieve the resolution of more sub-tasks).
Common Issues with Existing Bot Responses:

Failure to secure an acceptable price even after two negotiation attempts.
Poor ability to respond, leaving merchants' statements unaddressed.
Misunderstanding the price input by merchants.
Failing to suggest platform intervention when encountering self-adjustments.
Task Content:
Determine whether the current response by the bot requires improvement. If improvement is needed, write the optimized reply; if no optimization is needed, skip the annotation.
\end{Verbatim}
  \end{minipage}
\end{table*}

\end{document}